\documentclass{article}
\usepackage[utf8]{luainputenc}
\usepackage{array}
\usepackage{float}
\usepackage{booktabs}
\usepackage{textcomp}
\usepackage{multirow}
\usepackage{amsmath}
\usepackage{amssymb}
\usepackage{graphicx}

\makeatletter

\@ifundefined{pageheight}{\let\pageheight\pdfpageheight}{}
\@ifundefined{pagewidth}{\let\pagewidth\pdfpagewidth}{}
\pageheight\paperheight
\pagewidth\paperwidth

\providecommand{\tabularnewline}{\\}
\floatstyle{ruled}
\newfloat{algorithm}{tbp}{loa}
\providecommand{\algorithmname}{Algorithm}
\floatname{algorithm}{\protect\algorithmname}

\usepackage[utf8]{inputenc} 
\usepackage[T1]{fontenc}    
\usepackage{hyperref}       
\usepackage{url}            
\usepackage{booktabs}       
\usepackage{amsfonts}       
\usepackage{nicefrac}       
\usepackage{microtype}      
\usepackage[T1]{fontenc}
\usepackage{lmodern}

\usepackage{authblk}
\usepackage{hyperref}

\makeatother

\begin{document}

\author{}

\title{Wavelet Decomposition of Gradient Boosting}
\maketitle

\author{Shai Dekel,
  \and
  Oren Elisha,
  \and
  and Ohad Morgan
}

\newcommand{\Addresses}{{
  \bigskip
  \footnotesize

  Shai Dekel, \textsc{is with WIX AI and the School of Mathematics, Tel-Aviv
University, Tel-Aviv. }\nopagebreak
  \textit{shaidekel6@gmail.com }

  \medskip

  Oren Elisha, \textsc{is with Microsoft Israel and the School of Mathematics, Tel-Aviv
University, Tel-Aviv. }\nopagebreak
  \textit{orenelis@gmail.com }

  \medskip

  Ohad Morgan, \textsc{is with the School of Mathematics, Tel-Aviv
University, Tel-Aviv. }\nopagebreak
  \textit{ohadmorgan1989@gmail.com }

}}

\begin{abstract}
In this paper we introduce a significant improvement to the popular
tree-based Stochastic Gradient Boosting algorithm using a wavelet
decomposition of the trees. This approach is based on harmonic analysis
and approximation theoretical elements, and as we show through extensive
experimentation, our wavelet based method generally outperforms existing
methods, particularly in difficult scenarios of class unbalance and
mislabeling in the training data. 
\end{abstract}

\section{Introduction}

In the setting of regression and classification tasks on structured
data, decision tree ensembles are extremely useful as off-the-shelf
tools \cite{Hastie2009}, being relatively fast to construct, adaptive,
and when trees are small, they also produce interpretable models.
Tree-based boosting methods \cite{Hastie2009,Friedman00greedyfunction,xu2016shrinkage}
are popular ensemble methods that are constructed by a sequential
generation of pruned decision trees. These 'weak learners' are then
combined to form a 'strong' estimator. In this work we focus on improving
the well-known Gradient Boosting (GB) technique \cite{Friedman00greedyfunction,Friedman:2002:SGB:635939.635941},
which computes a sequence of trees, that are trained to predict the
residual between the response variable and the negative gradient direction
of the previous tree. These residuals are used as the response variables
for the next tree in the iterative processes so that the sum of the
weighted trees in the ensemble is the final estimator. 

Wavelets \cite{daubechies1992TenLecturesOnWavelets,mallat2008waveletSP}
are a powerful, yet simple, tool for sparse representations of 'complex'
functions. In \cite{dubossarsky2016waveletGB}, a wavelet-based GB
method was introduced, using a classic wavelet decomposition of the
original predictors, followed by a componentwise linear least squares
GB mechanism. Our wavelets approach is very different. We use the
theoretical foundation developed in \cite{ElishaandDekel2016WaveletDofRF}
and apply a wavelet decomposition of the decision trees formed during
the boosting process. This mathematical model allows us to design
a more robust pruning algorithm by re-ordering of the nodes of the
trees based on their significance. As we will show in the experimental
part, the adaptive wavelet pruning approach generally outperforms
existing methods, particularly in difficult scenarios of class unbalance
and mislabeling in the training data. In addition, we also demonstrate
how to employ our method on the stochastic GB version, which uses
bagging to gain diversity and use the Out Of Bag (OOB) samples to
improve generalization.

The rest of paper is organized as follows. In section 2, we overview
Gradient Boosting (GB) algorithms and in particular tree-based GB
algorithms. In addition, we describe the stochastic GB algorithm which
combines a bagging procedure into the GB. In section 3, we present
the “Geometric Wavelets” (GW) decomposition, and highlight some theoretical
and practical properties that emphasis their correspondence with sparsity.
In Section 4 we present the Geometric Wavelets Gradient Boosting (GWGB)
algorithm which combines stochastic GB with the GW decomposition.
In section 5, we conclude with experiment results that compare our
algorithm with competing boosting algorithms in different challenging
settings. 

\section{Gradient boosting trees\label{sec:Gradient-Boosting}}

\subsection{Decision trees \label{subsec: Decision-trees}}

In the setting of statistics and machine learning \cite{Hastie2009,breiman1984classificationAndRegressionTrees}
the construction we present in this chapter is referred as Decision
Tree or the Classification and Regression Tree (CART).

When we are given with a real-valued function or a discrete data-set
\begin{equation}
\left\{ x_{i}\in\Omega_{0},y_{i}=f(x_{i})\right\} _{i=1}^{m},\label{eq: discrete dataset}
\end{equation}
in some convex bounded domain $\Omega_{0}\subset\mathbb{R}^{n}$,
our goal is to find an efficient representation of this data $\hat{f}(x)$,
overcoming the complexity, geometry and possibly non-smooth nature
of the function values. The efficiency of $\hat{f}(x)$ is typically
estimated by minimization of a loss function $L$, with respect to
the data \eqref{eq: discrete dataset}, and is usually combined with
additional regularization condition that aims to reduce the generalization
error. For example as seen in \eqref{eq: Pruning minimizing condition},
a sparsity condition is applied to reduce overfitting artifacts.

The decision tree's first level is formed by a partition of the initial
domain $\Omega_{0}$, into two sub-domains, e.g. by intersecting it
with a hyper-plane, so as to minimize a given cost function. This
subdivision process then continues recursively on the nested sub-domains
until some exit criterion is met, which in turn, determines the leaves
of the tree. We now describe one instance of the cost function. At
each stage of the subdivision process, at a certain node of the tree,
the algorithm finds, for the convex domain $\Omega\subset\mathbb{R}^{n}$
associated with the node, a partition by an hyper-plane into two convex
sub-domains $\Omega',\Omega''$, and two multivariate low-order polynomials
$Q_{\Omega'},Q_{\Omega''}$, of fixed (typically low) total degree
$r-1$, that minimize the following quantity
\begin{equation}
\left\Vert f-Q_{\Omega'}\right\Vert _{L_{p(\Omega')}}^{p}+\left\Vert f-Q_{\Omega''}\right\Vert _{L_{p(\Omega'')}}^{p},\,\,\Omega'\cup\Omega''=\Omega.\label{eq: quantity to be minimized - decision trees}
\end{equation}
If the data-set is discrete, consisting of feature vectors $x_{i}\in\mathbb{R}^{n}$
, with response values $f(x_{i})$, then a discrete functional is
minimized
\begin{equation}
\sum_{x_{i}\in\Omega'}|f(x_{i})-Q_{\Omega'}(x_{i})|^{p}+\sum_{x_{i}\in\Omega''}|f(x_{i})-Q_{\Omega''}(x_{i})|^{p}
\end{equation}
Observe that for any given subdividing hyperplane, the approximating
polynomials in \eqref{eq: quantity to be minimized - decision trees}
can be uniquely determined for $p=2$, by least square minimization. 

For $r=1$ , the approximating polynomials are nothing but the mean
of the function values over each of the sub-domains
\begin{align}
Q_{\Omega'} & =c_{\Omega'}=\frac{1}{\#\{x_{i}\in\Omega'\}}\sum_{x_{i}\in\Omega'}f(x_{i}),\nonumber \\
Q_{\Omega''} & =c_{\Omega''}=\frac{1}{\#\{x_{i}\in\Omega''\}}\sum_{x_{i}\in\Omega''}f(x_{i}).\label{eq: constant app.}
\end{align}
Denoting by $\Omega_{t}^{j}$ a node on level j of the tree with counting
index $t$. It is easy to see that for each fixed level $J$, $\Omega_{0}=\bigcup_{t=1}^{2^{J}}\Omega_{t}^{J}$.
Therefore, we can describe the tree evaluated at any fixed level $J$
by $T^{J}(x)=\sum_{t=1}^{2^{J}}Q_{\Omega_{t}^{J}}(x)1_{\Omega_{t}^{J}}(x)$,
or simply by $T(x)$, when evaluation is done on the terminal nodes
or to a predefined level $J$. Here, $1_{\Omega}(x)=1$, if $x\in\Omega$
and $1_{\Omega}(x)=0$, if $x\notin\Omega$.

In classification problems, the input training set consists of labeled
data using $P$ classes instead of function values. In this scenario,
each input training point $x_{i}\in\mathbb{R}^{n}$ is assigned with
a class $C(x_{i})$. To convert the problem to the same ‘functional’
setting described above one assigns to each class $C$ the value of
a node on the regular simplex consisting of $P$ vertices in $\mathbb{R}^{P-1}$
(all with equal pairwise distances). Thus, we may assume that the
input data is in the form $\left\{ x_{i},y_{i}\right\} _{i=1}^{m}\in\left(\mathbb{R}^{n},\mathbb{R}^{P-1}\right).$
In this case, if we choose approximation using constants $(r=1)$,
then the calculated mean over any sub-domain $\Omega$ is in fact
a point $\overrightarrow{E}_{\Omega}\in\mathbb{R}^{P-1}$, inside
the simplex. Obviously, any value inside the multidimensional simplex,
can be mapped back to a class, along with an estimated certainty confidence
level, by calculating the closest vertex of the simplex to it. As
will become obvious, these mappings can be applied to any wavelet
approximation of functions receiving multidimensional values in the
simplex.

In many algorithms that are based on decision trees, the high-dimensionality
of the data does not allow to search through all possible subdivisions.
As in our experimental results, one may restrict the subdivisions
to the class of hyperplanes aligned with the main axes. In contrast,
there are cases where one would like to consider more advanced form
of subdivisions, where they take certain hyper-surface form, such
as conic-sections. Our paradigm of wavelet decomposition can support
in principle all of these forms.

\subsection{Pruning decision trees}

In many cases, the response variable $f(x_{i})$ in \eqref{eq: discrete dataset}
is obtained with noise of different types. Thus, bias-variance consideration
(e.g. avoiding over fit) \cite{Hastie2009} encourages pruning techniques
that restrict the size of a tree in varies ways. Pre-pruning \cite{kotsiantis2013DecisionTrees:aRecentOverview}
involves a “termination condition” to determine when it is desirable
to terminate some of the branches prematurely when a decision tree
is generated. For example, in \cite{Breiman:2001:RF:570181.570182}
a minimal node size is used as an exit criterion for tree generation.
On the other hand, post-pruning \cite{kotsiantis2013DecisionTrees:aRecentOverview},
may be applied to remove some of the branches after tree is generated
and could be evaluated. For example, in the CART algorithm \cite{breiman1984classificationAndRegressionTrees}
after a tree model had been generated, one applies a regularization
condition with a factor $\gamma$, that penalize adding more nodes,
by minimizing 
\begin{equation}
\left\Vert \sum_{i=1}^{m}\left(y_{i}-T(x_{i})\right)^{2}\right\Vert _{l_{2}}+\gamma\#\left\{ \Omega_{t}^{j}\in T(x),\,j=1,..J\right\} .\label{eq: Pruning minimizing condition}
\end{equation}
As will be described in the next sections, while most Boosting algorithms
\cite{Hastie2009} set a fixed level $J$ as a pruning strategy, our
GWGB method uses a different pruning approach that is based on the
data encapsulated in each node rather than the level of the tree. 

\subsection{Gradient boosting}

Gradient Boosting (GB) \cite{Friedman00greedyfunction,Mason99boostingalgorithms,Hastie2009}
is based on computing a sequence of weak learners such as pruned decision
trees, using a gradient descent iterative method. A functional gradient
view of boosting was first presented in \cite{Mason99boostingalgorithms}.
This led to the development of boosting algorithms in many areas of
machine learning and statistics, beyond regression and classification.

Let $X\in\mathbb{R}^{n}$ denote a real value random input vector,
and $Y\in\mathbb{R}^{P-1}$ a real value random output vector, with
$Pr(X,Y)$ their joint distribution. in such case we have a typically
unknown function $f(X)=E[Y|X]$ , so we may seek for an approximation
function $\hat{f}:\mathbb{R}^{n}\rightarrow\mathbb{R}^{P-1}$ based
on the training samples \eqref{eq: discrete dataset}. Ideally, the
approximation $\hat{f}(x)$ should be the one that minimizes the expected
prediction error with respect to some specified loss function $L(Y,f(X))$,
\begin{equation}
\hat{f}(X)=\underset{f}{argmin}E_{X,Y}L(Y,f(X)).\label{eq: expected prediction error}
\end{equation}
Frequently employed loss functions $L$ include squared-error $\left(y-f(x)\right){}^{2}$
and absolute error $\left|y-f(x)\right|$ for $y\in\mathbb{R}$ (regression),
and negative binomial log-likelihood $log\left(1+e^{-2yf(x)}\right)$,
when $y\in\left\{ -1,1\right\} $ (binary classification). Here, we
present a unified approach for regression and classification and choose
the squared-error, as described in \eqref{eq: quantity to be minimized - decision trees}
and \ref{sec: Geometric-wavelets}.

In the setting of tree-based GB algorithm, the approximation function
$\hat{f}(x)$ is a combination of $K+1$ weak learners 
\begin{equation}
\hat{f}_{K}(x):=\hat{f}_{0}(x)+\nu\sum_{k=1}^{K}T_{k}(x),\label{eq: f^on gradient tree boosting}
\end{equation}
where $K$ is the number of boosting iterations, $T_{k}(x)$ are the
pruned trees, and $\nu$ is the step size (also called Learning Rate
or Shrinkage).

The function $\hat{f}_{0}(x)$ is typically an initial estimate such
as 
\[
\hat{f}_{0}(x)=\underset{c}{argmin}\sum_{i=1}^{N}L\left(y_{i},c\right).
\]
To generate the weak learners, one typically constructs $K$ decision
trees $\left\{ T_{k}\right\} _{k=1}^{K}$, while applying a fixed
tree level $J$, in the following way. In each iteration $k$, a decision
tree $T_{k}$ is built, so residuals could be set as the response
variables for the next $k+1$ step 
\[
\left\{ y_{i}^{k+1}=\hat{f}_{k}(x_{i})-y_{i}^{k}\right\} _{i=1}^{m}\,\,\text{with}\,\,\left\{ y_{i}^{0}=y_{i}\right\} _{i=1}^{m}.
\]
 This iterative procedure resembles a gradient decent in the sense
that at each iteration we set the next step in the opposite direction
of the pseudo gradient of the loss function $L$.

As described in \cite{Hastie2009}, it is common practice to set $4\leq J\leq8$,
to have good results in the context of boosting. As we shall see,
our approach is to choose a higher level $J$ and then apply the wavelet-based
approach of pruning specific nodes. The selection of $K$ should set
a balance between reducing the training error and avoiding overfitting
when $K\rightarrow\infty$. Thus, \cite{R_Packege-gb} uses a validation
samples or an OOB samples (when applying the stochastic version of
GB), to find the optimal $K$. 

A modification to tree-based GB algorithm, called Stochastic GB (SGB),
was proposed in \cite{Friedman:2002:SGB:635939.635941} by applying
a bagging step at each iteration, that uses only a random subsample
of the training data. This randomly selected subsample is then used,
instead of the full sample, to fit the regression tree. The OOB samples
are then used for validation. In our algorithm, as well as previous
algorithms, the OOB are used to determine the pruning, not only validation.

\section{Geometric wavelets\label{sec: Geometric-wavelets}}

The Geometric Wavelet (GW) decomposition of decision trees were presented
at \cite{Dekel05adaptivemultivariate}, based on the theory of \cite{Karaivanov_nonlinearpiecewise,Karaivanov02algorithmsfor}.
It was recently generalized to a decomposition of Random Forests \cite{ElishaandDekel2016WaveletDofRF}
and used to enhance their performance as well as introduce a novel
algorithm for feature importance.

Let $\Omega'$ be a child of $\Omega$ in a decision tree $T$, i.e.
$\Omega'\subset\Omega$ and $\Omega'$ has been created by a partition
of $\Omega$, and let two polynomials $Q_{\Omega'}$ ,$Q_{\Omega''}$
that minimize the quantity \eqref{eq: quantity to be minimized - decision trees}.
We use the polynomial approximations $Q_{\Omega'}$,$Q_{\Omega}\in\Pi_{r-1}\left(\mathbb{R}^{n}\right)$
and define 
\begin{equation}
\psi_{\Omega'}:=\psi_{\Omega'}(f):=1_{\Omega'}(Q_{\Omega'}-Q_{\Omega})\label{eq: geometric wavelet}
\end{equation}
as the \textbf{geometric wavelet }associated with the sub-domain $\Omega'$
and the function $f$, or the given discrete data-set \eqref{eq: discrete dataset}.

Each wavelet $\psi_{\Omega'}$ is a ‘local difference’ component that
belongs to the detail space between two levels in the tree, a ‘low
resolution’ level associated with $\Omega$ and a ‘high resolution’
level associated with $\Omega'$. Also, the wavelets \eqref{eq: geometric wavelet}
have the ‘zero moments’ property, i.e., if the response variable is
sampled from a polynomial of degree $r-1$ over $\Omega$, then our
local scheme will compute $Q_{\Omega'}=Q_{\Omega}=f(x)$, $\forall x\in\Omega'$
and therefore $\psi_{\Omega'}=0$. 

Under certain mild conditions on the tree $T$ and the function $f$,
we have by the nature of the wavelets, the ‘telescopic’ sum of differences:
\begin{equation}
f=\sum_{\Omega\in T}\psi_{\Omega},\;\qquad\text{where}\;\;\psi_{\Omega_{0}}:=Q_{\Omega_{0}}.\label{eq: telescopic wavelet sum}
\end{equation}
For example, \eqref{eq: telescopic wavelet sum} holds in $L_{p}$-sense
($1\leq p<\infty$), if $f\in L_{p}(\Omega_{0})$, and for any $x\in\Omega_{0}$
and series of domains $\Omega_{j}\in T$, each on a level $j$ with
$x\in\Omega_{j}$, we have that $\underset{j\rightarrow\infty}{lim}diam(\Omega_{j})=0$
(see Theorem 2.1 in \cite{Dekel05adaptivemultivariate}).

In the theoretical setting, the norm of a wavelet is computed by
\begin{equation}
\left\Vert \psi_{\Omega'}\right\Vert _{2}^{2}=\int_{\Omega'}\left(Q_{\Omega'}(x)-Q_{\Omega}(x)\right)^{2}dx,\label{eq: wavelet norm}
\end{equation}
and in the discrete case by
\begin{equation}
\left\Vert \psi_{\Omega'}\right\Vert _{2}^{2}=\sum_{x_{i}\in\Omega'}\left|Q_{\Omega'}(x_{i})-Q_{\Omega}(x_{i})\right|{}^{2},\label{eq: wavelet norm discrete case}
\end{equation}
where $\Omega'$ is a child of $\Omega$. This wavelet norm tell us
how much information this wavelet encapsulate (see \cite{Dekel05adaptivemultivariate},\cite{ElishaandDekel2016WaveletDofRF}). 

Recall that our approach converts classification problems into a ‘functional’
setting by assigning the $P$ class labels to vertices of a simplex
in $\mathbb{R}^{P-1}$ (see discussion in \ref{subsec: Decision-trees})
. In such cases of multi-valued functions, choosing $r=1$, the wavelet
$\psi_{\Omega'}:\mathbb{R}^{n}\rightarrow\mathbb{R}^{P-1}$ is 
\[
\psi_{\Omega'}=1_{\Omega'}\left(\overrightarrow{E}_{\Omega'}-\overrightarrow{E}_{\Omega}\right),
\]
and its norm is given by 
\[
\left\Vert \psi_{\Omega'}\right\Vert _{2}^{2}=\left\Vert \overrightarrow{E}_{\Omega'}-\overrightarrow{E}_{\Omega}\right\Vert _{l_{2}}^{2}\#\left\{ x_{i}\in\Omega'\right\} ,
\]
where for $\overrightarrow{v}\in\mathbb{R}^{P-1}$, $\left\Vert \overrightarrow{v}\right\Vert :=\sqrt{\Sigma_{i=1}^{P-1}v_{i}^{2}}.$
Accordingly, we can consider the squared-error as loss function for
classification problems.

It is easy to see that the decision tree $T$ can be writen as 
\begin{equation}
T(x)=\sum_{\Omega_{j}\in T}\psi_{\Omega_{j}}(x).\label{eq: Wavelet form fo T}
\end{equation}
The theory (see Theorem 4 in \cite{Dekel05adaptivemultivariate})
tells us that sparse approximation is achieved by ordering the wavelet
components based on their norm
\begin{equation}
\left\Vert \psi_{\Omega_{k_{1}}}\right\Vert _{2}\geq\left\Vert \psi_{\Omega_{k_{2}}}\right\Vert _{2}\geq\left\Vert \psi_{\Omega_{k_{3}}}\right\Vert _{2}...\label{eq: ordering the wavelet components based on their norm}
\end{equation}
Thus, the adaptive $M$-term approximation of a decision tree $T$
is
\begin{equation}
T_{M}(x):=\sum_{j=1}^{M}\psi_{\Omega_{k_{j}}}.\label{eq:M-term approximation}
\end{equation}
This pruning method is, in some sense, a generalization of the classical
$M$-term wavelet sum, where the wavelets are constructed over dyadic
cubes (see \cite{DeVore98nonlinearapproximation}).

\section{Wavelet decomposition of gradient boosting}

In this section, we introduce a combination of Stochastic GB Tree
algorithm and Geometric Wavelet decomposition. In our setting, instead
of the pruned decision tree $T$ at some fixed level $J$ as weak
learner, we retrieve the $M$ ``most important'' nodes, in term
of wavelet-norm (see \eqref{eq: ordering the wavelet components based on their norm}
and \eqref{eq:M-term approximation}), which is the $M$-term approximation
$T_{M}$. This $M$-term approximation is used as the weak learner
at each boosting iteration. To select $M$ at each iteration, we use
the OOB data as in \ref{sec:Gradient-Boosting} and add the wavelets,
one by one according to its wavelets norm until error of the model
is minimized on the OOB set. Our implementation for ``Geometric Wavelets
Gradient Boosting'' described in Algorithm \ref{alg: Geometric Wavelets Gradient Boosting}.
\begin{algorithm}[h]
\caption{\label{alg: Geometric Wavelets Gradient Boosting}Geometric Wavelets
Gradient Boosting }

\begin{enumerate}
\item \begin{raggedright}
Initialize $\hat{f}_{0}(x)=\underset{c}{argmin}\sum_{i=1}^{m}L\left(y_{i},c\right).$
\par\end{raggedright}
\item \begin{raggedright}
set $\left\{ y_{i}^{0}=y_{i}\right\} _{i=1}^{m}$.
\par\end{raggedright}
\item \begin{raggedright}
For $k=1,2,...,K$ 
\par\end{raggedright}
\begin{enumerate}
\item \begin{raggedright}
update the residuals $\left\{ y_{i}^{k}=\hat{f}_{k-1}(x_{i})-y_{i}\right\} _{i=1}^{m}$.
\par\end{raggedright}
\item \begin{raggedright}
Choose randomly subset of $m'$ variables from the original data-set,
denote by $\left\{ x_{i,}y_{i}^{k}\right\} _{i=1}^{m'}$. based on
this training set generate a tree $T(x)=\sum_{j}\psi_{\Omega_{k_{j}}}$,
where $\left\{ \psi_{\Omega_{k_{j}}}\right\} _{j}$ are the GW sorted
by wavelet norm (see $\ref{eq: ordering the wavelet components based on their norm}$).
\par\end{raggedright}
\item \begin{raggedright}
Denote the OOB subset by $OOB=\left\{ \left(x,y\right)|\:\left(x,y\right)\notin\left\{ x_{i,}y_{i}\right\} _{i=1}^{m'}\right\} ,$
then compute:
\[
M_{k}=\underset{M}{argmin}\sum_{(x,y)\in OOB}L\left(y^{k},\sum_{j=1}^{M}\psi_{\Omega_{k_{j}}}(x)\right).
\]
\par\end{raggedright}
\item \begin{raggedright}
Update the prediction model: $\hat{f}_{k}(x)=\hat{f}_{k-1}(x)+\nu\sum_{j=1}^{M_{k}}\psi_{\Omega_{k_{j}}}(x)$.
\par\end{raggedright}
\end{enumerate}
\item \raggedright{}Output $\hat{f}(x)=\hat{f}_{K}(x)$.
\end{enumerate}
\end{algorithm}
. 

\begin{figure}
\begin{centering}
\includegraphics[scale=0.75]{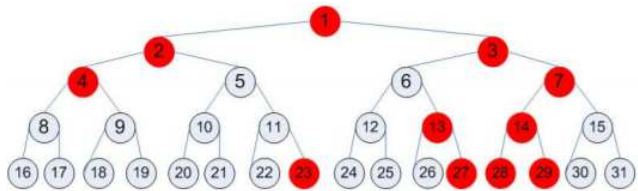}
\par\end{centering}
\caption{Illustration of greedy node selection by wavelet norms\label{fig: greedy node selection by wavelet norms-1}}
\end{figure}

One of the advantages of this form of tree pruning is the fact that
nodes are selected according to their contribution to the prediction
\eqref{eq: telescopic wavelet sum}, rather than their position at
a certain level in the tree. This allows an adaptive selection of
high and low resolution at the same step of the boosting. An illustration
of $M$-term GW collection whose graph representation includes some
unconnected components is shown in Figure \ref{fig: greedy node selection by wavelet norms-1}.
The $M$-term nodes are marked in red, while the rest of the nodes
in the tree are not used for the estimation.

Another advantage of relaying on the wavelets norms for pruning in
the GBM setting, is an efficient feature selection for the ensemble.
In some cases, explanatory attributes may be non-descriptive and even
noisy, leading to the creation of problematic nodes in the decision
trees. Nevertheless, in these cases, the corresponding wavelet norms
are controlled and these nodes can be omitted from the representation
\eqref{eq:M-term approximation}. An example that demonstrates this
phenomen is presented in \cite{ElishaandDekel2016WaveletDofRF} (see
Example 1). The example shows that with high probability, the wavelets
associated with the correct variables have relatively higher norms
than wavelets associated with non-descriptive variables. Hence the
wavelet based criterion will choose, with high probability the correct
variable. Since the tree partitions are based on \eqref{eq: quantity to be minimized - decision trees}
and \eqref{eq: ordering the wavelet components based on their norm},
the non-descriptive variables are less likely to form partitions that
are part of the GWGB ensemble.

\section{Experimental results}

In this section we compare the Geometric Wavelets GB algorithm (GWGB,
algorithm \ref{alg: Geometric Wavelets Gradient Boosting}) with other
boosting and bagging methods in terms of classifying imbalance datasets,
improving regression tasks and overcoming mislabeling noise in classification
tasks.

Our GWGB code was written in C\# and is publicly available \footnote{https://github.com/ohadmorgan/GeometricWaveletGradeintBoosting.git}.
At each iteration we use the OOB technique as described in section
\ref{sec:Gradient-Boosting}, with 80\% of the training set to build
the wavelets tree, and 20\% for $M$-term selection. Moreover, a fixed
step size ($\nu$) of 0.1 is used throughout all of our experiments. 

\subsection{Classification with imbalance class distributions \label{subsec:Imbalance-Classification-problem}}

Classification problem with data-sets that suffer from imbalanced
class distributions is a challenging problem in the field of machine
learning. 
\begin{table}
\begin{centering}
\caption{Class Imbalance results comparison (AUC)\label{tab:Class-Imbalance-Results}}
\par\end{centering}
\centering{}%
\begin{tabular}{>{\raggedright}p{2cm}>{\centering}p{1cm}>{\centering}p{1cm}>{\centering}p{1cm}>{\centering}p{1cm}}
\toprule 
\multirow{2}{2cm}{\textbf{\scriptsize{}Dataset name}} & \textbf{\scriptsize{}Best Bagging-based method} & \textbf{\scriptsize{}Best Boosting-based method } & \textbf{\scriptsize{}Best Classic method} & \textbf{\scriptsize{}Geometric Wavelets}\tabularnewline
 & {\scriptsize{}UB4} & {\scriptsize{}RUS 1} & {\scriptsize{}SMT} & {\scriptsize{}GWGB}\tabularnewline
\midrule
\textbf{\scriptsize{}glass1} & {\scriptsize{}0.737} & {\scriptsize{}0.763} & {\scriptsize{}0.737} & \textbf{\scriptsize{}0.816}\tabularnewline
\textbf{\scriptsize{}ecoli0vsl} & {\scriptsize{}0.980} & {\scriptsize{}0.969} & {\scriptsize{}0.973} & \textbf{\scriptsize{}0.986}\tabularnewline
\textbf{\scriptsize{}Wisconsin} & {\scriptsize{}0.960} & {\scriptsize{}0.964} & {\scriptsize{}0.953} & \textbf{\scriptsize{}0.985}\tabularnewline
\textbf{\scriptsize{}Pima} & {\scriptsize{}0.760} & {\scriptsize{}0.726} & {\scriptsize{}0.725} & \textbf{\scriptsize{}0.809}\tabularnewline
\textbf{\scriptsize{}Iris0} & {\scriptsize{}0.990} & {\scriptsize{}0.990} & {\scriptsize{}0.990} & \textbf{\scriptsize{}1.000}\tabularnewline
\textbf{\scriptsize{}glass0} & {\scriptsize{}0.814} & {\scriptsize{}0.813} & {\scriptsize{}0.775} & \textbf{\scriptsize{}0.880}\tabularnewline
\textbf{\scriptsize{}yeast1} & {\scriptsize{}0.722} & {\scriptsize{}0.719} & {\scriptsize{}0.709} & \textbf{\scriptsize{}0.775}\tabularnewline
\textbf{\scriptsize{}vehicle1} & {\scriptsize{}0.787} & {\scriptsize{}0.747} & {\scriptsize{}0.730} & \textbf{\scriptsize{}0.810}\tabularnewline
\textbf{\scriptsize{}vehicle2} & {\scriptsize{}0.964} & {\scriptsize{}0.970} & {\scriptsize{}0.950} & \textbf{\scriptsize{}0.982}\tabularnewline
\textbf{\scriptsize{}vehicle3} & {\scriptsize{}0.802} & {\scriptsize{}0.765} & {\scriptsize{}0.728} & \textbf{\scriptsize{}0.805}\tabularnewline
\textbf{\scriptsize{}Haberman} & \textbf{\scriptsize{}0.664} & {\scriptsize{}0.655} & {\scriptsize{}0.616} & {\scriptsize{}0.651}\tabularnewline
\textbf{\scriptsize{}glass0123vs456} & {\scriptsize{}0.904} & {\scriptsize{}0.930} & {\scriptsize{}0.923} & \textbf{\scriptsize{}0.960}\tabularnewline
\textbf{\scriptsize{}vehicle0} & {\scriptsize{}0.952} & {\scriptsize{}0.958} & {\scriptsize{}0.919} & \textbf{\scriptsize{}0.982}\tabularnewline
\textbf{\scriptsize{}ecoli1} & {\scriptsize{}0.900} & {\scriptsize{}0.883} & {\scriptsize{}0.911} & \textbf{\scriptsize{}0.951}\tabularnewline
\textbf{\scriptsize{}new-thyroid2} & {\scriptsize{}0.958} & {\scriptsize{}0.938} & {\scriptsize{}0.966} & \textbf{\scriptsize{}0.996}\tabularnewline
\textbf{\scriptsize{}new-thyroid1} & {\scriptsize{}0.964} & {\scriptsize{}0.958} & {\scriptsize{}0.963} & \textbf{\scriptsize{}0.993}\tabularnewline
\textbf{\scriptsize{}ecoli2} & {\scriptsize{}0.884} & {\scriptsize{}0.899} & {\scriptsize{}0.811} & \textbf{\scriptsize{}0.918}\tabularnewline
\textbf{\scriptsize{}Segimmt0} & {\scriptsize{}0.988} & \textbf{\scriptsize{}0.993} & \textbf{\scriptsize{}0.993} & {\scriptsize{}0.987}\tabularnewline
\textbf{\scriptsize{}glass6} & {\scriptsize{}0.904} & {\scriptsize{}0.918} & {\scriptsize{}0.884} & \textbf{\scriptsize{}0.935}\tabularnewline
\textbf{\scriptsize{}yeast3} & {\scriptsize{}0.934} & {\scriptsize{}0.925} & {\scriptsize{}0.891} & \textbf{\scriptsize{}0.957}\tabularnewline
\textbf{\scriptsize{}ecoli3} & {\scriptsize{}0.908} & {\scriptsize{}0.856} & {\scriptsize{}0.812} & \textbf{\scriptsize{}0.923}\tabularnewline
\textbf{\scriptsize{}Page-blocks0} & {\scriptsize{}0.958} & {\scriptsize{}0.948} & {\scriptsize{}0.950} & \textbf{\scriptsize{}0.990}\tabularnewline
\textbf{\scriptsize{}yeast2vs4} & {\scriptsize{}0.936} & {\scriptsize{}0.933} & {\scriptsize{}0.859} & \textbf{\scriptsize{}0.981}\tabularnewline
\textbf{\scriptsize{}yeast05679vs4} & {\scriptsize{}0.794} & {\scriptsize{}0.803} & {\scriptsize{}0.760} & \textbf{\scriptsize{}0.863}\tabularnewline
\textbf{\scriptsize{}vowel0} & {\scriptsize{}0.947} & {\scriptsize{}0.943} & {\scriptsize{}0.951} & \textbf{\scriptsize{}0.988}\tabularnewline
\textbf{\scriptsize{}glass016vs2} & \textbf{\scriptsize{}0.754} & {\scriptsize{}0.617} & {\scriptsize{}0.606} & {\scriptsize{}0.720}\tabularnewline
\textbf{\scriptsize{}glass2} & {\scriptsize{}0.769} & \textbf{\scriptsize{}0.780} & {\scriptsize{}0.639} & {\scriptsize{}0.690}\tabularnewline
\textbf{\scriptsize{}ecoli4} & {\scriptsize{}0.888} & \textbf{\scriptsize{}0.942} & {\scriptsize{}0.779} & {\scriptsize{}0.906}\tabularnewline
\textbf{\scriptsize{}suttle0vs4} & \textbf{\scriptsize{}1.000} & \textbf{\scriptsize{}1.000} & \textbf{\scriptsize{}1.000} & \textbf{\scriptsize{}1.000}\tabularnewline
\textbf{\scriptsize{}yrast1vs7} & \textbf{\scriptsize{}0.786} & {\scriptsize{}0.715} & {\scriptsize{}0.700} & {\scriptsize{}0.760}\tabularnewline
\textbf{\scriptsize{}glass4} & {\scriptsize{}0.846} & {\scriptsize{}0.915} & {\scriptsize{}0.887} & \textbf{\scriptsize{}0.963}\tabularnewline
\textbf{\scriptsize{}page-blocks13vs4} & {\scriptsize{}0.978} & {\scriptsize{}0.987} & \textbf{\scriptsize{}0.996} & {\scriptsize{}0.992}\tabularnewline
\textbf{\scriptsize{}abalone9vs18} & {\scriptsize{}0.719} & {\scriptsize{}0.693} & {\scriptsize{}0.628} & \textbf{\scriptsize{}0.827}\tabularnewline
\textbf{\scriptsize{}glass016vs5} & {\scriptsize{}0.943} & \textbf{\scriptsize{}0.989} & {\scriptsize{}0.813} & {\scriptsize{}0.946}\tabularnewline
\textbf{\scriptsize{}suttle2vs4} & \textbf{\scriptsize{}1.000} & \textbf{\scriptsize{}1.000} & {\scriptsize{}0.992} & {\scriptsize{}0.994}\tabularnewline
\textbf{\scriptsize{}yrast1458vs7} & \textbf{\scriptsize{}0.606} & {\scriptsize{}0.567} & {\scriptsize{}0.537} & {\scriptsize{}0.594}\tabularnewline
\textbf{\scriptsize{}glass5} & {\scriptsize{}0.949} & {\scriptsize{}0.943} & {\scriptsize{}0.881} & \textbf{\scriptsize{}0.982}\tabularnewline
\textbf{\scriptsize{}yeast2vs8} & {\scriptsize{}0.783} & {\scriptsize{}0.789} & \textbf{\scriptsize{}0.834} & {\scriptsize{}0.616}\tabularnewline
\textbf{\scriptsize{}yeast4} & {\scriptsize{}0.855} & {\scriptsize{}0.812} & {\scriptsize{}0.712} & \textbf{\scriptsize{}0.865}\tabularnewline
\textbf{\scriptsize{}yeast1289vs7} & {\scriptsize{}0.734} & {\scriptsize{}0.721} & {\scriptsize{}0.683} & \textbf{\scriptsize{}0.765}\tabularnewline
\textbf{\scriptsize{}yeast5} & {\scriptsize{}0.952} & {\scriptsize{}0.959} & {\scriptsize{}0.934} & \textbf{\scriptsize{}0.968}\tabularnewline
\textbf{\scriptsize{}ecoli0137vs26} & {\scriptsize{}0.745} & {\scriptsize{}0.794} & \textbf{\scriptsize{}0.814} & \textbf{\scriptsize{}0.814}\tabularnewline
\textbf{\scriptsize{}yeast6} & {\scriptsize{}0.869} & {\scriptsize{}0.823} & {\scriptsize{}0.829} & \textbf{\scriptsize{}0.876}\tabularnewline
\textbf{\scriptsize{}Abalone19} & \textbf{\scriptsize{}0.721} & {\scriptsize{}0.631} & {\scriptsize{}0.521} & {\scriptsize{}0.594}\tabularnewline
\bottomrule
\end{tabular}
\end{table}

We present a comparison of our algorithm performance, with state of
the art ensemble-based techniques for imbalanced data-sets presented
in \cite{galar2012reviewEnsemblesClassImbalance}. 
\begin{table*}
\caption{Regression results comparison\label{tab:Regression-Results-Comparison}}

\centering{}%
\begin{tabular}{>{\centering}p{1cm}>{\centering}p{1cm}>{\centering}p{1cm}>{\centering}p{1cm}>{\centering}p{1cm}>{\centering}p{1cm}>{\centering}p{1cm}>{\centering}p{1cm}}
\textbf{\scriptsize{}Dataset } & \multicolumn{3}{>{\centering}p{3cm}}{{\scriptsize{}Decision stumps}} & \multicolumn{3}{>{\centering}p{3cm}}{{\scriptsize{}Vanilla neural networks}} & \multirow{2}{1cm}{\textbf{\scriptsize{}GWGB}}\tabularnewline
\cmidrule{1-7} 
\textbf{\scriptsize{}name} & \textbf{\scriptsize{}R-}{\scriptsize \par}

\textbf{\scriptsize{}Boosting} & \textbf{\scriptsize{}$\epsilon$-Boosting} & \textbf{\scriptsize{}RT-}{\scriptsize \par}

\textbf{\scriptsize{}Boosting} & \textbf{\scriptsize{}R-}{\scriptsize \par}

\textbf{\scriptsize{}Boosting} & \textbf{\scriptsize{}$\epsilon$-Boosting} & \textbf{\scriptsize{}RT-}{\scriptsize \par}

\textbf{\scriptsize{}Boosting} & \tabularnewline
\midrule
\textbf{\scriptsize{}Diabetes} & {\scriptsize{}58.71\textpm 1.2} & {\scriptsize{}58.94\textpm 1.9} & {\scriptsize{}58.61\textpm 2.7} & {\scriptsize{}58.03\textpm 2.3} & {\scriptsize{}58.03\textpm 2.4} & {\scriptsize{}58.03\textpm 2.4} & \textbf{\scriptsize{}57.01\textpm 3.0}\tabularnewline
\midrule
\textbf{\scriptsize{}Housing} & {\scriptsize{}4.13\textpm 0.3} & {\scriptsize{}4.33\textpm 0.2} & {\scriptsize{}4.14\textpm 0.4} & {\scriptsize{}4.02\textpm 0.3} & {\scriptsize{}4.45\textpm 0.3} & {\scriptsize{}4.45\textpm 0.3} & \textbf{\scriptsize{}3.38\textpm 0.4}\tabularnewline
\midrule
\textbf{\scriptsize{}CCS} & {\scriptsize{}5.47\textpm 0.1} & {\scriptsize{}6.10\textpm 0.7} & {\scriptsize{}5.35\textpm 0.3} & {\scriptsize{}6.59\textpm 0.4} & {\scriptsize{}6.62\textpm 0.2} & {\scriptsize{}6.52\textpm 0.3} & \textbf{\scriptsize{}4.80\textpm 0.4}\tabularnewline
\midrule
\textbf{\scriptsize{}Abalone} & {\scriptsize{}2.28\textpm 0.02} & {\scriptsize{}2.40\textpm 0.05} & {\scriptsize{}2.28\textpm 0.05} & {\scriptsize{}2.12\textpm 0.05} & \textbf{\scriptsize{}2.10\textpm 0.03} & {\scriptsize{}2.13\textpm 0.02} & {\scriptsize{}2.17\textpm 0.06}\tabularnewline
\bottomrule
\end{tabular}
\end{table*}
\begin{table*}
\caption{\label{tab:Classification-Results-Compariso}Misclassification results
comparison }

\centering{}%
\begin{tabular}{>{\centering}p{1cm}c>{\centering}p{1cm}>{\centering}p{1cm}>{\centering}p{1cm}>{\centering}p{1cm}>{\centering}p{1cm}>{\centering}p{1cm}}
\textbf{\scriptsize{}Dataset name} & \textbf{\scriptsize{}NL} & \textbf{\scriptsize{}r-Ada Boost} & \textbf{\scriptsize{}rBoost-Fixed $\gamma$} & \textbf{\scriptsize{}rBoost} & \textbf{\scriptsize{}GBoost} & \textbf{\scriptsize{}MBoost} & \textbf{\scriptsize{}GWGB}\tabularnewline
\midrule
\multirow{2}{1cm}{\textbf{\scriptsize{}Banana}} & \textbf{\scriptsize{}0.1} & {\scriptsize{}86.87\textpm 1.1} & {\scriptsize{}87.06\textpm 0.9} & {\scriptsize{}87.04\textpm 0.9} & {\scriptsize{}83.91\textpm 1.6} & {\scriptsize{}78.13\textpm 3.4} & \textbf{\scriptsize{}87.60\textpm 1.6}\tabularnewline
\cmidrule{2-8} 
 & \textbf{\scriptsize{}0.3} & {\scriptsize{}85.27\textpm 3.0} & \textbf{\scriptsize{}85.53\textpm 2.1} & {\scriptsize{}85.06\textpm 2.7} & {\scriptsize{}79.38\textpm 1.6} & {\scriptsize{}75.31\textpm 2.5} & {\scriptsize{}85.49\textpm 1.3}\tabularnewline
\midrule
\multirow{2}{1cm}{\textbf{\scriptsize{}PID}} & \textbf{\scriptsize{}0.1} & {\scriptsize{}74.20\textpm 2.3} & {\scriptsize{}74.37\textpm 1.5} & {\scriptsize{}74.80\textpm 2.4} & {\scriptsize{}72.60\textpm 2.0} & \textbf{\scriptsize{}75.67\textpm 1.9} & {\scriptsize{}74.21\textpm 6.2}\tabularnewline
\cmidrule{2-8} 
 & \textbf{\scriptsize{}0.3} & {\scriptsize{}72.53\textpm 1.9} & {\scriptsize{}70.43\textpm 2.4} & {\scriptsize{}71.43\textpm 2.3} & {\scriptsize{}69.40\textpm 2.9} & {\scriptsize{}73.33\textpm 2.3} & \textbf{\scriptsize{}75.65\textpm 7.5}\tabularnewline
\midrule
\multirow{2}{1cm}{\textbf{\scriptsize{}Heart}} & \textbf{\scriptsize{}0.1} & {\scriptsize{}78.40\textpm 3.1} & {\scriptsize{}79.70\textpm 3.5} & {\scriptsize{}79.10\textpm 4.4} & {\scriptsize{}76.40\textpm 3.1} & {\scriptsize{}77.60\textpm 3.5} & \textbf{\scriptsize{}80.74\textpm 8.2}\tabularnewline
\cmidrule{2-8} 
 & \textbf{\scriptsize{}0.3} & \textbf{\scriptsize{}78.50\textpm 4.0} & {\scriptsize{}77.40\textpm 6.5} & {\scriptsize{}78.10\textpm 4.3} & {\scriptsize{}70.00\textpm 5.5} & {\scriptsize{}75.20\textpm 3.7} & {\scriptsize{}73.70\textpm 11.5}\tabularnewline
\midrule
\multirow{2}{1cm}{\textbf{\scriptsize{}Two-Norm}} & \textbf{\scriptsize{}0.1} & {\scriptsize{}95.70\textpm 0.8} & {\scriptsize{}95.58\textpm 0.9} & {\scriptsize{}95.59\textpm 0.7} & {\scriptsize{}90.35\textpm 1.0} & {\scriptsize{}92.79\textpm 0.5} & \textbf{\scriptsize{}96.40\textpm 0.8}\tabularnewline
\cmidrule{2-8} 
 & \textbf{\scriptsize{}0.3} & {\scriptsize{}93.33\textpm 0.9} & {\scriptsize{}93.13\textpm 1.3} & {\scriptsize{}93.40\textpm 1.1} & {\scriptsize{}83.94\textpm 2.0} & {\scriptsize{}91.16\textpm 0.9} & \textbf{\scriptsize{}94.82\textpm 0.7}\tabularnewline
\bottomrule
\end{tabular}
\end{table*}
The experiment is based on the testing a variety of methodologies
on 44 real-world imbalanced problems from KEEL data-set repository
\cite{alcala2010keel}. We use the same 5-fold cross-validation data
and partitions that are provided in \cite{galar2012reviewEnsemblesClassImbalance}
to measure the AUC (Area Under the Curve) metric. Moreover, at each
iteration we grow the tree to a fixed level of depth level $8$, and
selected $M_{k}$ terms according to our algorithm, while using the
same $K=10$ which was used in \cite{galar2012reviewEnsemblesClassImbalance}.
Comparison results, including GWGB algorithm, are presented in table
\ref{tab:Class-Imbalance-Results}. Since the authors of \cite{galar2012reviewEnsemblesClassImbalance}
reviewed 37 different bagging, boosting and classics algorithms, for
brevity and space limitation, we present a comparison of our method
to the best algorithm in each category (in term of mean AUC) for each
data set.The success of our method is due to the fact that we could
build deeper trees and reach high resolution areas where the rare
categories might accord, and select these nodes in early stages of
the ensemble (lower $K$). This is the advantage of wavelets reordering
according to thier norm which enables a pruning strategy that is not
dependent on the tree's depth or level. 

\subsection{Regression}

In this section we compare our method with the most recent Boosting
schemes presented at \cite{xu2016shrinkage}. We follow the same randomization
process presented by \cite{xu2016shrinkage} with 20 random trails
of 2-fold cross validation, and we have followed the same technique
for adaptive selection of the number of iterations as in \cite{xu2016shrinkage}
as we chose the best $k\in[0,500]$ in term of RMSE on validation.
As in the previous section, at each iteration we have grow the tree
to a fixed level of depth level $8$, and select $M_{k}$ terms according
to our algorithm. The results are presented in Table \ref{tab:Regression-Results-Comparison}
are the average RMSE and standard deviation of the 20 random trails
and are compared to the three best algorithms from \cite{xu2016shrinkage}.

\subsection{Overcoming mislabeling noise in classification}

Boosting methods are known to be sensitive to label noise \cite{bootkrajang2013boosting}.
The experiment is based on the same testing methodology presented
in \cite{bootkrajang2013boosting}, with the injection of two noise
levels (NL) of random 10\% and 30\% of the original labels in the
datasets\@. Average and standard deviation of the misclassification
rates computed from 10-fold cross validation. As in \cite{bootkrajang2013boosting}
we have restricted the number of iteration to $K=150$, and restricted
the level to$2$.

The first method (rAdaBoost) is modification of the AdaBoost algorithm,
using \char`\"{}robust classifiers\char`\"{} that combined and boosted
using known AdaBoost algorithm. The two next methods (rBoost-Fixed
gamma and rBoost) are new robust boosting algorithms where the objective
function is a convex combination of two exponential losses. The results
for the well-known Gentle Boost (GBoost), and Modest Boost (MBoost)
are also taken from \cite{bootkrajang2013boosting} and presented
in Table \ref{tab:Classification-Results-Compariso}. The reason for
improved results of our method in the case of mislabeling noise is
due to removing wavelets with small GW norm. As seen in \ref{eq: wavelet norm discrete case}
the magnitude of a wavelet norm that corresponds to a single point
(typical miss labeled point) is small and hence will be typically
pruned while keeping informative nodes on the same level of the tree.

\bibliographystyle{plain}
\bibliography{WGB}

\Addresses

\end{document}